\documentclass[letterpaper,journal]{IEEEtran}
\usepackage{amsmath,amsfonts}
\usepackage{algorithmic}
\usepackage{algorithm}
\usepackage{array}
\usepackage[caption=false,font=normalsize,labelfont=sf,textfont=sf]{subfig}
\usepackage{textcomp}
\usepackage{stfloats}
\usepackage{url}
\usepackage{verbatim}
\usepackage{graphicx}
\usepackage{cite}
\usepackage{hyperref}
\begin{document}

\title{Depth-Aware Implicit Neural Representation Priors for 3D Gravity Inversion}

\author{\uppercase{León Suarez-Rodriguez}, 
\uppercase{Paul Goyes-Pe\~nafiel}, \uppercase{Javier Torres-Quintero}, AND \uppercase{Henry Arguello} \thanks{Department of Systems Engineering and Informatics, Universidad Industrial de Santander, Colombia, 680002}}


\maketitle

\begin{abstract}
Gravimetry images subsurface density contrasts associated with geological structures, geothermal systems, and intrusive bodies. Recovering a three-dimensional density model from gravity observations is highly ill-posed because of its non-uniqueness, limited data coverage, and the attenuation of the gravity field with depth. Classical inversion methods rely on explicit regularization and parameter tuning, whereas supervised deep-learning approaches require representative gravity--density pairs that are rarely available. This paper proposes an unsupervised depth-aware implicit neural representation for 3D gravity inversion. The density volume is represented by multiple coordinate-based neural networks assigned to overlapping depth slabs and optimized directly from the observed gravity measurements through the sensitivity matrix. Slab-specific Fourier features, physics-based depth gains, and scheduled regularization provide structural priors without requiring labeled density models. Experiments on four synthetic scenarios show that the proposed method provides better overall performance in terms of RMSE, PSNR, and SSIM than the evaluated conventional and neural baselines. It also recovers more compact and spatially coherent density bodies, improves the separation of nearby anomalies, preserves internal structures, and reconstructs their vertical extent better. These results indicate that the proposed depth-aware formulation helps to mitigate the depth ambiguity inherent in gravity inversion. In the field experiment, where no ground-truth density model was available, the method produced compact, separated, and vertically coherent anomalies consistent with the observed gravity pattern.
\end{abstract}

\begin{IEEEkeywords}
gravity inversion, implicit neural representation, Fourier features, depth weighting, inverse problems
\end{IEEEkeywords}

\section{Introduction}

\IEEEPARstart{G}{ravity} data are widely used in subsurface exploration because they provide indirect information regarding density variations at depth. This information is relevant to several geophysical applications, including mineral exploration, geological mapping, and geothermal energy assessment \cite{nabighian2005historical,hinze2013gravity, 9743443, 9703360}. In geothermal exploration, gravity anomalies can help identify density contrasts associated with intrusive bodies, hydrothermal alteration zones, and volcanic conduits, which are commonly linked to fluid circulation and heat transport \cite{maithya2020analysis, chen2023saltGravity}. However, estimating the 3D subsurface density distribution from surface gravity measurements is a highly ill-posed inverse problem owing to non-uniqueness, limited data coverage, and the natural attenuation of the gravity field with depth \cite{parker1975theory, 9703360}. Therefore, improving the accuracy and reliability of gravity inversion is essential, because erroneous density estimates can lead to inaccurate geological interpretations, misplaced exploration targets, and increased costs during subsequent drilling or development stages \cite{hinze2013gravity, 10341312}.

Classical gravity inversion methods commonly formulate the problem as an iterative optimization task, where the density model is recovered by minimizing a data-misfit term together with regularization terms that impose prior assumptions on the solution \cite{li19983}. These approaches include smoothness-constrained inversion, compact inversion, focusing regularization, sparsity-promoting methods, and model-based iterative schemes \cite{portniaguine1999focusing,vatankhah20173,last1983compact}. Although these techniques have been successfully applied in many scenarios, their performance strongly depends on the choice of regularization, parameter tuning, discretization strategy, and the initial model. Moreover, the resulting density distribution may be overly smooth, structurally biased, or unable to recover complex geological geometries when the imposed prior does not adequately represent the true subsurface structure.

On the other hand, deep learning methods have been widely explored as data-driven alternatives for solving inverse problems. In a supervised learning setting, a neural network is trained on input-output pairs to approximate a nonlinear mapping from the measured data to the target model \cite{ongie2020deep,arridge2019solving}. After training, the learned model can provide rapid predictions and may capture complex statistical relationships that are difficult to describe using explicit analytical priors. In gravity inversion, supervised deep learning methods have been used to map gravity anomaly data directly to subsurface density distributions, commonly using convolutional or encoder-decoder architectures trained on synthetic gravity-density pairs \cite{huang2021deep,li2023fast,chen2024three,9325520,9546806,ggac190,Zhang2022, 10225600, 10110924}. Although these approaches have shown promising reconstruction capabilities and fast inference, they require representative training datasets with known ground-truth density models, that are rarely available in geophysical applications. Furthermore, synthetic training data may introduce a domain gap, reducing the reliability of the inversion when applied to field measurements with different noise levels, geological conditions, acquisition geometries, or physical assumptions \cite{10613855, 9847099, 9782500, 11311505, XU2022104685, TorresQuintero2025}.

To alleviate the dependence on labeled density models, unsupervised and self-supervised deep learning strategies have been proposed for gravity inversion. These methods embed the sensitivity matrix into the training process, allowing the predicted density distribution to be optimized through the misfit between simulated and observed gravity responses rather than through direct supervision from ground-truth models \cite{Li2022Self3DGravity,Wu2023PhysicalConstraint,Zhou2024SelfConstrained, 10839036, Li2026ImplicitInversion}. Additional constraints, such as depth weighting, have been incorporated to improve depth resolution and data consistency in geothermal applications \cite{Zhou2024DepthWeighting}. 

To address these limitations, this paper proposes an unsupervised depth-aware implicit neural representation for 3D gravity inversion. In the proposed formulation, the subsurface density distribution was represented by multiple coordinate-based neural networks \cite{3495724.3496350,3495724.3496356,8953655,10167682}, each associated with a slab-shaped subvolume of the inversion domain. The final 3D density cube was reconstructed by merging the slab-wise predictions during optimization. Unlike supervised learning approaches, the proposed method does not require paired gravity-density training data; instead, it is optimized directly from the observed gravity measurements through the sensitivity matrix. This physics-consistent formulation combines the flexibility of implicit neural representations with an unsupervised inversion strategy, enabling volumetric density reconstruction while reducing dependence on labeled datasets and fixed voxel-wise model parameterizations. This per-instance optimization follows the broader principle of untrained neural priors for inverse problems, in which the network parametrization itself acts as an implicit regularizer \cite{ulyanov2018a, 3495724.3496350,3495724.3496356,8953655,10167682}.
\begin{figure*}[!t]
        \centering
    \includegraphics[width=1\linewidth]{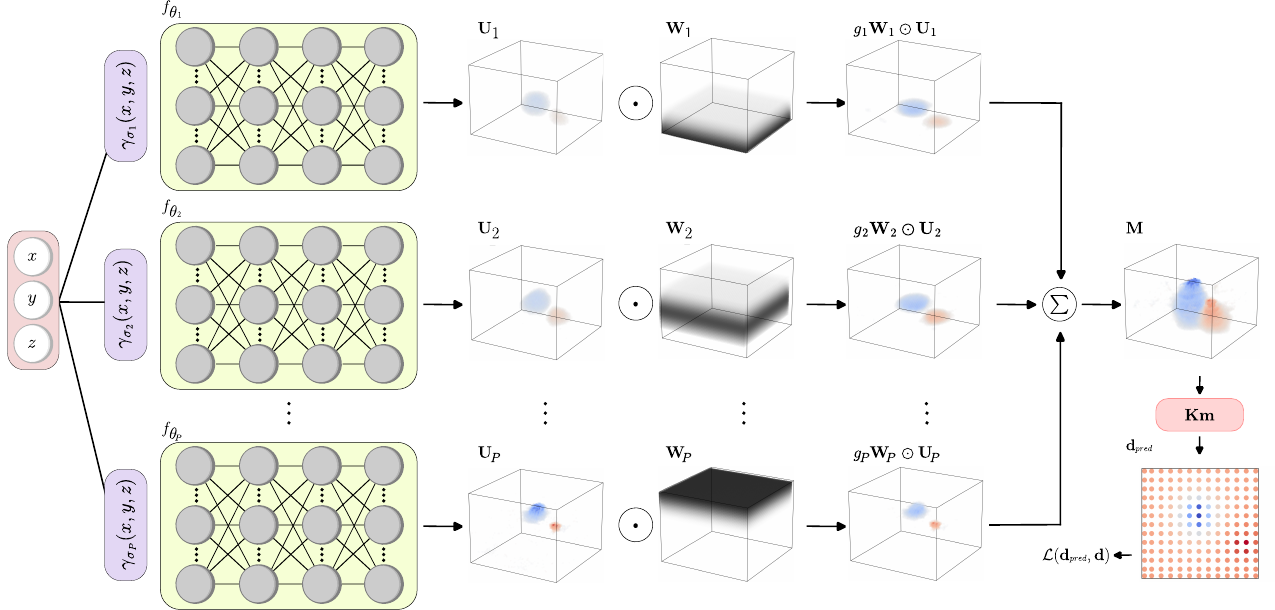}
    \caption{Overview of the proposed Depth-Aware Implicit Neural Representation reconstruction. Model input coordinates $(x,y,z)$ are first mapped through a Fourier-feature encoding $\gamma(x,y,z)$ and then evaluated by $P$ independent multilayer perceptrons $\{f_{\boldsymbol{\theta}_i}\}$, each associated with a Gaussian slab window $\mathbf{W}_i$. Each network predicts a volumetric component $\mathbf{U}_i$, which is masked by its corresponding slab window through the Hadamard product. The final reconstructed volume is obtained by summing all slab-wise contributions weighted by $g_i$. The sensitivity matrix $\mathbf{K}$ is then applied to $\mathrm{vec}(\mathbf{M})$ to generate the estimated data $d_{\mathrm{pred}}$, and the network parameters are optimized by minimizing the objective function between the observed data and the estimated measurements.
}
\label{fig:slabwise_reconstruction}
\end{figure*}

The main contributions of this work are summarized as follows:
\begin{itemize}
\item We introduce a depth-aware implicit neural parameterization for 3D gravity inversion. The density volume is represented as a partition-of-unity combination of $P$ coordinate-based neural networks associated with overlapping depth slabs. Each network is assigned a slab-specific Fourier-feature bandwidth and a fixed physics-informed depth gain, enabling the representation capacity to vary according to the depth-dependent resolution of gravity data.

\item We developed a self-supervised inversion framework in which the gravity sensitivity matrix was embedded directly into the objective function. The optimization combines scheduled total-variation, sparsity, and inter-slab consensus regularization: the structural priors gradually decrease, while the consensus penalty increases to promote agreement among neighboring slab representations and produce a spatially coherent density model.

\item We evaluated the proposed method using synthetic and field gravity data. In the synthetic experiments, it achieved better overall reconstruction metrics and recovered more compact, spatially coherent structures with improved body separation and depth localization compared to the baselines. The field experiment produces compact and vertically coherent density anomalies that preserve the main spatial pattern of the observed gravity data, while providing a more localized interpretation than the baseline reconstructions.

\end{itemize}

\section{Method}

The proposed method takes the model-grid coordinates as input, lifts them using slab-specific Fourier-feature encodings $\gamma_{\sigma_i}(\cdot)$ \cite{3495724.3496356}, and passes the resulting features through $P$ independent coordinate networks ${f_{\boldsymbol{\theta}_i}}$. The network outputs $\mathbf{U}_i$ are confined to overlapping depth bands by Gaussian slab windows $\mathbf{W}_i$, scaled by the physics-based depth gains $g_i$, and fused to form the reconstructed density volume $\hat{\mathbf{M}}$. The sensitivity matrix $\mathbf{K}$ then maps the vectorized volume to the predicted response $\mathbf{d}_{\mathrm{pred}}$. The network parameters are optimized by minimizing the data-misfit term together with the scheduled regularization terms. The following subsections describe each component in detail, while Figure \ref{fig:slabwise_reconstruction} provides an overview of the proposed inversion framework.

\subsection{Problem overview}
 
Let $\mathbf{m}=\operatorname{vec}(\mathbf{M}) \in \mathbb{R}^{N_m}$ denote the vectorized three-dimensional density-contrast model $\mathbf{M}\in\mathbb{R}^{X\times Y\times Z}$, where $X$, $Y$, and $Z$ denote the number of grid cells along the two horizontal directions and the vertical direction, respectively, and $N_m=X \cdot Y \cdot Z$. The gravity observations acquired at
$N_d$ receiver locations are collected in the vector $\mathbf{d}\in\mathbb{R}^{N_d}$. The subsurface domain is discretized into $N_m$ homogeneous rectangular prisms. The vertical gravitational response of each prism at each receiver location is computed using the analytical rectangular-prism formulation of Nagy \cite{Nagy1966}. The resulting responses are assembled into the sensitivity matrix $\mathbf{K}\in\mathbb{R}^{N_d\times N_m}$, whose element $K_{ij}$ represents the gravity response at the $i$-th receiver produced by a unit density contrast in the $j$-th prism. The forward problem is then expressed as:

\begin{equation}
\mathbf{d} = \mathbf{K}\mathbf{m},
\end{equation}

Since $N_m \gg N_d$, estimating $\mathbf{m}$ from $\mathbf{d}$ is an ill-posed and underdetermined inverse problem, allowing many equivalent density-contrast models to fit the measurements. Additionally, gravimetry suffers from depth ambiguity: the sensitivity matrix decays with depth, meaning that a shallow, low-density body and a deeper, higher-density body can produce the same surface measurements. Resolving this depth ambiguity is therefore a central challenge in gravimetric inversion.

\begin{table*}[!t]
\caption{Hyperparameter configuration of the proposed method and the IGI baseline for the synthetic experiments. S1--S3 denote the hollow, syncline, and two-body benchmarks shown in Figs.~\ref{fig:synthetic3d} and \ref{fig:syntheticslice}. The same IGI architecture is used in all four cases, with the grid and number of observations set according to the corresponding experiment.}
\label{tab:params_synthetic}
\centering
\resizebox{\textwidth}{!}{%
\begin{tabular}{l|c|c|c|c|c}
\hline
& \multicolumn{4}{c|}{\textbf{Proposed method}}
& \textbf{IGI baseline} \\
\textbf{Parameter}
& \textbf{Cerro Mach\'in}
& \textbf{S1 hollow}
& \textbf{S2 syncline}
& \textbf{S3 two bodies}
& \textbf{All cases} \\
\hline
Grid $(X\times Y\times Z)$
& $50\times50\times50$
& $32\times32\times16$
& $32\times32\times16$
& $32\times32\times16$
& Corresponding case \\
Observations $N_d$
& 676 & 1024 & 1024 & 1024
& Corresponding case \\
Slabs $P$
& 14 & 12 & 12 & 14
& --- \\
Window factor $\kappa$
& 0.7 & 0.7 & 0.7 & 0.7
& --- \\
Fourier frequencies $L$
& 64 & 64 & 64 & 64
& --- \\
Max. bandwidth $\sigma_{\max}$
& 48 & 10.06 & 10.06 & 48
& --- \\
Hidden width
& 128 & 128 & 128 & 128
& \begin{tabular}{c}
1024, 1024, 1024, 1024, 512,\\
256, 64, 16, 8
\end{tabular} \\
Hidden layers
& 4 & 5 & 5 & 4
& 9 \\
Depth exponent $\beta$
& 3.0 & 3.94 & 3.94 & 5.0
& --- \\
Depth offset $z_0$
& 3.0 & 2.0 & 2.0 & 3.0
& --- \\
$\lambda_{\mathrm{TV}}$
& 0.1
& $1.23\times10^{-2}$
& $1.23\times10^{-2}$
& 0.3
& --- \\
$\lambda_{L_1}$
& 0.1
& $1.86\times10^{-2}$
& $1.86\times10^{-2}$
& 0.03
& --- \\
$\lambda_{\mathrm{c}}$
& 24 & 31.27 & 31.27 & 24
& --- \\
Decay rate $r$
& 10 & 10 & 10 & 10
& --- \\
Learning rate
& $10^{-3}$
& $5\times10^{-4}$
& $5\times10^{-4}$
& $10^{-3}$
& $10^{-3}$ \\
Iterations $T$
& 3500 & 3500 & 3500 & 3500
& 5000 \\
\hline
\end{tabular}%
}
\end{table*}
 
\subsection{Slab parametrization}
\label{sec:method}
Instead of directly optimizing the full volume, $\mathbf{M} \in \mathbb{R}^{X \times Y \times Z}$, we decompose it into $P$ overlapping horizontal slabs stacked along the depth axis, with each slab generated by an independent coordinate network. The slab predictions are localized at their corresponding depths through fixed soft windows and combined to form the complete density volume. 

This decomposition is motivated by the depth-dependent sensitivity of gravity measurements. Surface gravity data become progressively less sensitive to density variations at greater depths; therefore, cells located at similar depths are constrained with comparable strength and spatial resolution, whereas cells at substantially different depths are not. Assigning a separate generator to each depth band allows the representation bandwidth and depth gain to be adapted to the physical resolution of that region. Accordingly, the density volume is expressed as follows:

\begin{equation}
    \mathbf{M}=\sum_{i=1}^P \mathbf{W}_i\odot \mathbf{U}_i,
\end{equation}
where $\mathbf{U}_i \in \mathbb{R}^{X\times Y\times Z}$ is the $i$-th slab field, a full-grid 3D density volume produced by a neural network $f_{\boldsymbol{\theta}_i}$; $\mathbf{W}_i \in \mathbb{R}^{X\times Y\times Z}$ is the depth window of slab $i$, a soft window that depends on the voxel depth index $z$, whose weights $\mathbf{W}_i[x,y,z]$ lie in $[0,1]$, are large near the slab's center depth, and taper to zero away from it; and $\odot$ denotes the Hadamard product.
 
The windows are overlapping Gaussians centered at $c_i = (i-1)\frac{Z-1}{P-1}$, where $Z$ is the total depth and $i = 1,\dots,P$ indexes the slabs. The normalized window value at voxel $[x,y,z]$ is:
\begin{equation}
    \mathbf{W}_i[x,y,z] = \frac{\phi_i(z)}{\sum_{j=1}^{P} \phi_j(z)}, \ 
    \phi_i(z) = \exp\!\Big(-\frac{(z-c_i)^2}{2\,\sigma_w^2}\Big),
\end{equation}
which is $\approx 1$ at the center depth $c_i$ and decays with distance from it. The width $\sigma_w = \kappa\, Z/P$ is a fraction $\kappa < 1$ of the slab spacing, so adjacent bumps overlap: a hard box window would produce seams, whereas the overlap enables consensus between adjacent slabs and a seamless blend between neighbors. By construction, the windows form a partition of unity, $\sum_{i=1}^{P} \mathbf{W}_i[x,y,z] = 1$.

\subsection{Positional encoding}  
Each $i$-th slab field is computed by a trained implicit neural representation (INR) with weights $\boldsymbol{\theta}_i$ as follows:
\begin{equation} \mathbf{U}_i \;=\; \big\{ \tanh\!\big(f_{\boldsymbol{\theta}_i}\!\big( \gamma_{\sigma_i}(\boldsymbol{r})\big)\big) \;\big|\; \boldsymbol{r} \in \boldsymbol{\mathcal{R}} \big\}, \label{eq:slabfield} 
\end{equation} 
where $\boldsymbol{\mathcal{R}}=\{\boldsymbol{r}_1, \boldsymbol{r}_2, \dots, \boldsymbol{r}_{N_m}\}$ is the set of all grid coordinates, $\boldsymbol{r}=(x,y,z)$ denotes an individual coordinate, and $\gamma_{\sigma_i}(\cdot)$ is a Fourier-feature positional encoding \cite{3495724.3496356} defined by $\gamma_{\sigma_i}(\boldsymbol{r})=[\cos (2\pi \mathbf{B}_i \boldsymbol{r}), \sin(2\pi \mathbf{B}_i \boldsymbol{r})] \in \mathbb{R}^{2L}$, with $\mathbf{B}_i \in \mathbb{R}^{L \times 3}$ a random Gaussian matrix whose entries are sampled from $\mathcal{N}(0, \sigma_i^2\mathbf{I})$ and $L$ the number of random Fourier frequencies. The encoding counteracts the spectral bias of coordinate MLPs, enabling the networks to learn high-frequency content during the optimization, and the $\tanh$ bounds each slab field to $(-1,1)$, so the slabs produce normalized density fields whose physical amplitude is set later by the depth gains and the density bound $\rho_{\max}$.  The bandwidth $\sigma_i$ controls the spectral content that each network can represent, and assigning one bandwidth per slab encodes the depth-dependent resolution of gravity data directly into the parametrization: since the sensitivity matrix increasingly low-passes the model with depth, deep slabs are restricted to smooth, low-frequency content that the data can actually constrain, while shallow slabs retain the capacity to express fine detail. Moreover, drawing an independent $\mathbf{B}_i$ for each slab decorrelates the slab generators, so the networks start with diverse, independent priors that are progressively tied together via the regularization term as shown in \eqref{eq:concensus}.

\subsection{Depth weighting}
 Surface gravity data constrain the depth of a source only weakly, so the prior must supply it. We assign each slab a fixed gain adapted from the depth-weighting regularization of \cite{Li1998}, which compensates the $(z+z_0)^{-2}$ decay of the sensitivity with depth.

The reference depth is anchored to the acquisition physics rather than to an assumed grid orientation. Let $\mathcal{I}_z$ denote the set of column indices of $\mathbf{K}$ associated with the active voxels of layer $z$. The average sensitivity of layer $z$ is:
\begin{equation}
s(z)=\frac{1}{|\mathcal{I}_z|}\sum_{j\in\mathcal{I}_z}\big\|\mathbf{K}[:,j]\big\|_2 ,
\label{eq:layer_sensitivity}
\end{equation}
and the surface-layer index is estimated as:
\begin{equation}
z_{\mathrm{surf}}=\arg\max_z s(z).
\label{eq:surface_layer}
\end{equation}
This is motivated by the fact that the gravitational response of a model cell decreases with its distance from the observation points, so the voxels closest to the receiver surface exhibit the largest average sensitivity across the sensor array.  Writing $b(z)=|z-z_{\mathrm{surf}}|$ for the vertical distance, in grid cells, between layer $z$ and the surface layer, the depth gain is:
\begin{equation}
    \operatorname{gain}(z)=\left(\frac{b(z)+z_0}{z_0}\right)^{\frac{\beta}{2}},
\end{equation}
 where the constant $z_0$ stabilizes the weighting near the surface and sets the transition scale, and $\beta$ is the depth-weighting exponent that compensates the sensitivity decay \cite{Li1998}. The gain assigned to each slab is then:
\begin{equation}
    g_i = \frac{\sum_z w_i (z) \operatorname{gain}(z)}{\sum_z w_i (z)}, \qquad
    g_i \leftarrow \frac{g_i}{\min_j g_j},
    \label{eq:gains}
\end{equation}
where $w_i (z)=\mathbf{W}_i[x,y,z]$ denotes the depth profile of the window, which is independent of $(x,y)$. The renormalization $\min_j g_j = 1$ anchors the gain of the most sensitive, surface-adjacent slab to one: the surface is where the sensitivity of $\mathbf{K}$ is highest and therefore requires no amplification, so this slab passes through at its native scale, while every other slab is boosted in proportion to its distance from the surface layer.

\subsection{Objective function}

We optimize the parameters of the set of INRs $\mathcal{F}_{\boldsymbol{\Theta}}=\{f_{\boldsymbol{\theta}_i}(\gamma_{\sigma_i}(\cdot))\}_{i=1}^{P}$, with $\boldsymbol{\Theta}=\{\boldsymbol{\theta}_i\}_{i=1}^{P}$. The final model is given by:
\begin{equation}
    \hat{\mathbf{M}}(\boldsymbol{\Theta})=\rho_{\max} \tanh\left(\sum_{i=1}^{P} g_i \mathbf{W}_i \odot \mathbf{U}_i \right).
\end{equation}
To optimize the model weights, we propose the following optimization problem:
\begin{equation}
\begin{aligned}
    \boldsymbol{\Theta}^*=\arg\min_{\boldsymbol{\Theta}} &\|  \mathbf{K} \operatorname{vec} (\hat{\mathbf{M}}(\boldsymbol{\Theta})) - \mathbf{d}\|_2^2 +  \lambda_{\mathrm{TV}}(t) \mathrm{TV}(\hat{\mathbf{M}}(\boldsymbol{\Theta}))\\&+\lambda_{L_1}(t) \|\hat{\mathbf{M}}(\boldsymbol{\Theta})\|_1+\lambda_{\mathrm{c}}(t) \mathcal{C}(\boldsymbol{\Theta}),
\end{aligned}
\end{equation}
after which the reconstruction is normalized as $\hat{\mathbf{M}}(\boldsymbol{\Theta})\leftarrow\hat{\mathbf{M}}(\boldsymbol{\Theta})/\rho_{\max}$.
 
We use the anisotropic total variation \cite{RUDIN1992259} as an $L_1$ penalty on the spatial gradient; it drives most gradient entries to exactly zero (flat interiors) while tolerating a few large ones (sharp jumps), thereby preserving edges and yielding piecewise-constant bodies:
\begin{equation}
\begin{aligned}
    \mathrm{TV}(\hat{\mathbf{M}}(\boldsymbol{\Theta})) =
\frac{1}{N_x}\sum_{x,y,z} \big|\tilde m[x{+}1,y,z] - \tilde m[x,y,z]\big| &\\ +
\frac{1}{N_y}\sum_{x,y,z} \big|\tilde m[x,y{+}1,z] - \tilde m[x,y,z]\big| &\\ +
\frac{1}{N_z}\sum_{x,y,z} \big|\tilde m[x,y,z{+}1] - \tilde m[x,y,z]\big|,
\end{aligned}
\end{equation}
where $\tilde m \equiv \hat{\mathbf{M}}(\boldsymbol{\Theta})$. The sparsity term $\| \hat{\mathbf{M}}(\boldsymbol{\Theta}) \|_1$, the sum of absolute densities, pushes the background toward exactly zero, so the reconstruction consists of a few compact anomalies in an empty volume rather than a diffuse haze.

\begin{figure*}[!t]
    \centering
    \includegraphics[width=1\linewidth]{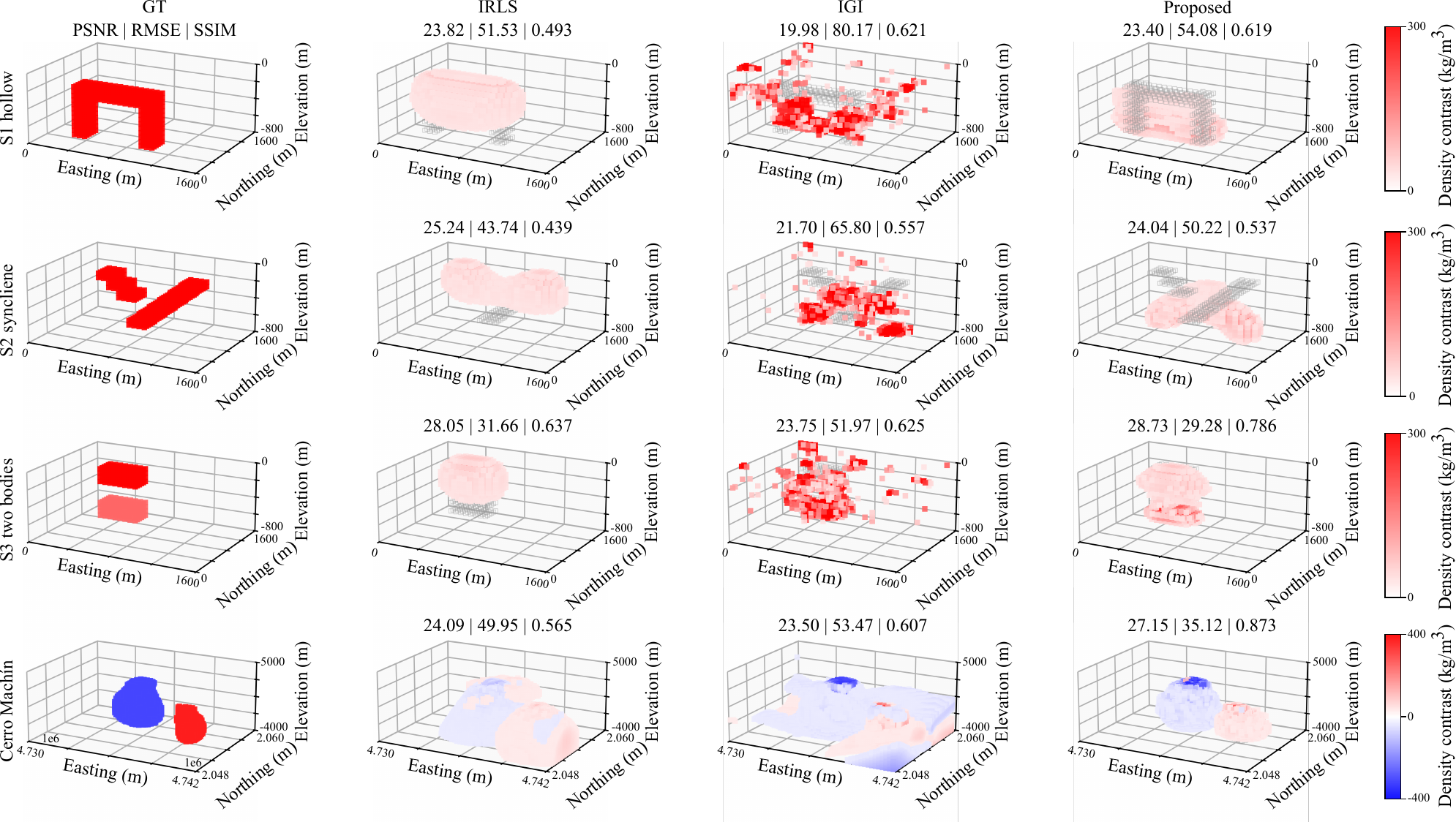}
    \caption{Three-dimensional comparison of the ground-truth density models (GT) with the reconstructions obtained by IRLS, IGI, and the proposed method for the S1 hollow, S2 syncline, S3 two-body, and Cerro Mach\'in synthetic scenarios. The values above each reconstruction are reported in the order PSNR (dB), RMSE ($\mathrm{kg\,m^{-3}}$), and SSIM. Higher PSNR and SSIM and lower RMSE indicate better model-domain agreement with the ground truth. Each row corresponds to one scenario, and all methods use the same density-contrast scale for that scenario.}
    \label{fig:synthetic3d}
\end{figure*}

Finally, we measure the disagreement between adjacent slabs as the mean squared difference of the raw slab fields $\mathbf{U}_i$ and $\mathbf{U}_{i+1}$, summed over neighboring pairs:
\begin{equation}
    \mathcal{C}(\boldsymbol{\Theta}) = \sum_{i=1}^{P-1} \frac{1}{N_m} \sum_{(x,y,z)\,\in\,\boldsymbol{\mathcal{R}}} \big(\mathbf{U}_i[x,y,z] - \mathbf{U}_{i+1}[x,y,z]\big)^2,
    \label{eq:concensus}
\end{equation}
where the inner sum is the mean squared difference between two slab fields over the $N_m$ voxels of the grid; a small $\mathcal{C}(\boldsymbol{\Theta})$ means the slabs describe one consistent field rather than $P$ contradictory ones.
 
The regularization strengths follow the schedules $\lambda_{\mathrm{TV}}(t) = \frac{\lambda_{\mathrm{TV}}}{1 + r\,t}$, $\lambda_{L_1}(t) = \frac{\lambda_{L_1}}{1 + r\,t}$, and $\lambda_{\mathrm{c}}(t) = \lambda_{\mathrm{c}}\,\frac{t}{T-1}$, where $\lambda_{\mathrm{TV}}$, $\lambda_{L_1}$, and $\lambda_{\mathrm{c}}$ are the base weights, $t$ is the iteration index, $r$ is a decay rate, and $T$ is the total number of iterations.

\section{Experiments and results}

\subsection{Baselines}
\label{sec:baselines}

The proposed method was compared with regularized and neural inversion approaches using the same observation vector, sensitivity matrix, computational grid, and active-cell configuration in each experiment. The synthetic comparison includes iteratively reweighted least squares (IRLS) \cite{101093} and implicit gravity inversion (IGI) \cite{Li2026ImplicitInversion}. IRLS represents a conventional regularized inversion strategy, whereas IGI is the closest neural baseline because it also optimizes an untrained coordinate network directly from the gravity observations. Unlike the proposed depth-aware formulation, however, IGI uses a single implicit representation for the complete density volume. The field-data comparison additionally includes the $L_0$-mixed and $L_1$-mixed methods of \cite{sadraeifar2026improved}. The $L_0$-mixed formulation promotes sharper and more localized models, while the $L_1$-mixed formulation favors smoother and more spatially continuous structures.

\subsection{Experiment I: Synthetic data}

Four controlled synthetic scenarios were considered: the S1 hollow-body, S2 syncline, and S3 two-body models, all adapted from \cite{Li2026ImplicitInversion}, and an irregular model designed to resemble the Cerro Mach\'in volcanic system. The gravity sensitivity matrix was constructed from the analytical vertical attraction of rectangular prisms using the formulation of Nagy \cite{Nagy1966}. Table~\ref{tab:params_synthetic} summarizes the inversion grids and optimization settings used in these experiments. The hyperparameters of the proposed method were selected independently for each synthetic case through a grid search using the model-domain Root Mean Squared Error (RMSE) with respect to the corresponding ground truth. All coordinate networks used SiLU activations \cite{ELFWING20183}; depending on the number of slabs and hidden layers, the complete compositional representation contains approximately one million trainable parameters.
\begin{figure*}[!t]
    \centering
    \includegraphics[width=1\linewidth]{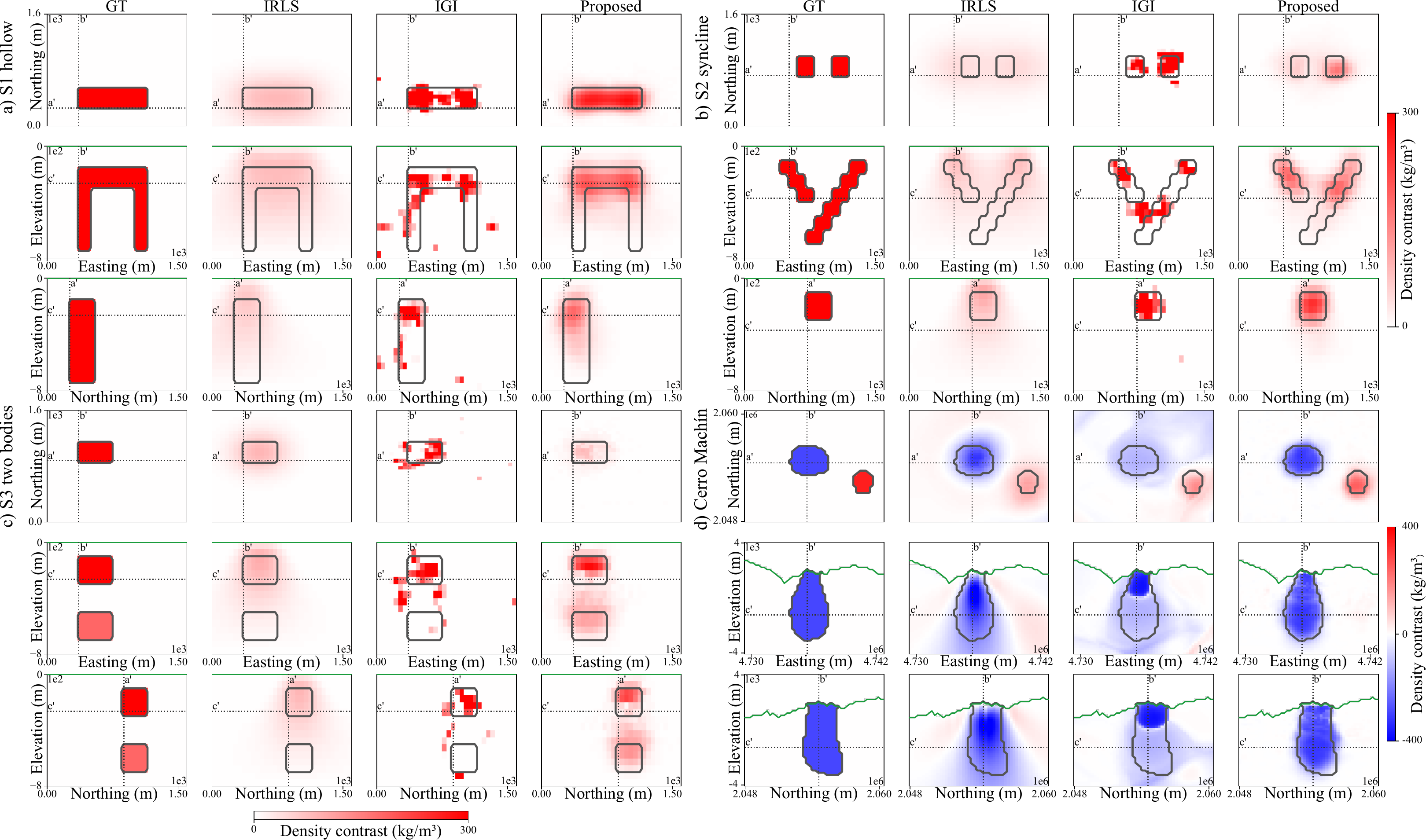}
    \caption{Cross-sectional comparison for the four synthetic scenarios: (a) S1 hollow, (b) S2 syncline, (c) S3 two bodies, and (d) Cerro Mach\'in. The columns show the ground truth (GT), IRLS, IGI, and the proposed method. The black contours mark the ground-truth support, and the green curve in the Cerro Mach\'in sections indicates the topographic surface.}
    \label{fig:syntheticslice}
\end{figure*}
Because the synthetic density models are known, reconstruction quality is evaluated directly in the model domain using the three complementary measures reported in Figure \ref{fig:synthetic3d}: peak signal-to-noise ratio (PSNR), RMSE, and structural similarity (SSIM) \cite{1284395}. The PSNR is expressed in decibels and increases as the reconstruction approaches the ground truth. The RMSE is expressed in $\mathrm{kg\,m^{-3}}$ and measures the voxel-wise density error, so lower values are better. The SSIM evaluates the similarity of the spatial structure and ranges toward one for increasingly similar volumes. These model-domain measures are interpreted together with the three-dimensional views and cross-sections because a single scalar does not fully describe cavity preservation, continuity, body separation, or depth extent. The receiver-domain RMSE, expressed in mGal, is analyzed separately from the model-domain metrics.

Figure~\ref{fig:synthetic3d} compares the ground-truth density volumes with the IRLS, IGI, and proposed reconstructions. Averaged over the four scenarios, the proposed method achieves a PSNR of $25.83$~dB, an RMSE of $42.18~\mathrm{kg\,m^{-3}}$, and an SSIM of $0.704$. The corresponding averages are $25.30$~dB, $44.22~\mathrm{kg\,m^{-3}}$, and $0.534$ for IRLS, and $22.23$~dB, $62.85~\mathrm{kg\,m^{-3}}$, and $0.603$ for IGI. The proposed representation therefore provides the best average RMSE and SSIM and a higher average PSNR than IRLS. The improvement is most pronounced for geometries that require separation of multiple sources or recovery of substantial vertical extent.

For S1, IRLS obtains the highest PSNR and lowest RMSE, with $23.82$~dB and $51.53~\mathrm{kg\,m^{-3}}$, while the proposed method obtains $23.40$~dB and $54.08~\mathrm{kg\,m^{-3}}$. IGI produces the largest voxel-wise error, with $19.98$~dB and $80.17~\mathrm{kg\,m^{-3}}$. In contrast, the structural comparison favors the neural representations: SSIM is $0.619$ for the proposed method and $0.621$ for IGI, compared with $0.493$ for IRLS. The proposed result offers the most useful balance between these measures because it preserves the hollow organization with substantially less fragmentation than IGI, although its walls remain broadened and the cavity is partially filled.

For S2, IRLS again gives the strongest voxel-wise scores, reaching $25.24$~dB PSNR and $43.74~\mathrm{kg\,m^{-3}}$ RMSE. The proposed method reaches $24.04$~dB and $50.22~\mathrm{kg\,m^{-3}}$, while IGI reaches $21.70$~dB and $65.80~\mathrm{kg\,m^{-3}}$. The SSIM values are $0.439$, $0.557$, and $0.537$ for IRLS, IGI, and the proposed method, respectively. Although IGI gives the highest SSIM by a small margin, its estimate is visibly fragmented. The proposed reconstruction better maintains the continuity of the two syncline limbs and their change in depth, providing a more coherent interpretation than the disconnected IGI response and a better-defined geometry than the smooth IRLS lobes.
\begin{figure}[!t]

    \centering
    \includegraphics[width=1\linewidth]{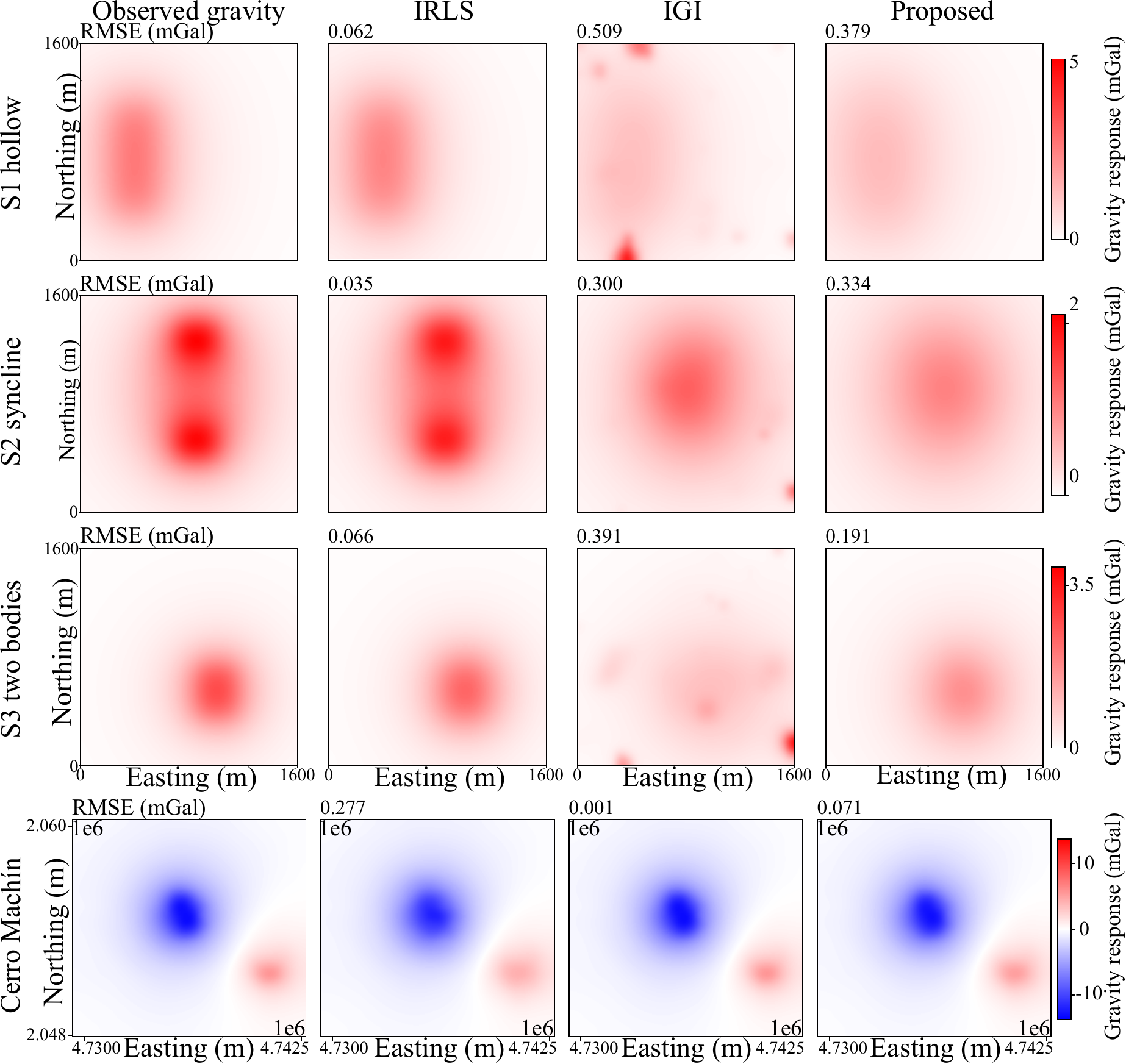}
    \caption{Observed gravity responses and responses predicted by IRLS, IGI, and the proposed method for the four synthetic scenarios. The number above each reconstruction is the receiver-domain RMSE in mGal.}
    \label{fig:responsesynth}
\end{figure}
The proposed method performs best according to all three metrics in S3. It obtains $28.73$~dB PSNR, $29.28~\mathrm{kg\,m^{-3}}$ RMSE, and $0.786$ SSIM. IRLS obtains $28.05$~dB, $31.66~\mathrm{kg\,m^{-3}}$, and $0.637$, whereas IGI obtains $23.75$~dB, $51.97~\mathrm{kg\,m^{-3}}$, and $0.625$. These quantitative gains agree with the clearer separation of the upper and lower bodies in the proposed reconstruction. IRLS merges the sources into a broader response, and IGI introduces isolated and discontinuous structures.

The largest improvement occurs in the Cerro Mach\'in scenario. The proposed method achieves $27.15$~dB PSNR, $35.12~\mathrm{kg\,m^{-3}}$ RMSE, and $0.873$ SSIM, outperforming IRLS at $24.09$~dB, $49.95~\mathrm{kg\,m^{-3}}$, and $0.565$, and IGI at $23.50$~dB, $53.47~\mathrm{kg\,m^{-3}}$, and $0.607$. Relative to IRLS, the proposed reconstruction increases PSNR by $3.06$~dB, reduces RMSE by $14.83~\mathrm{kg\,m^{-3}}$, and increases SSIM by $0.308$. It recovers the dominant negative, vertically elongated body together with the smaller positive anomaly to the east while preserving their relative locations and signs. Its boundaries remain smoother than the ground truth, but the model is substantially more faithful in both voxel-wise accuracy and structural organization.

\begin{figure}
    \centering
    \includegraphics[width=1\linewidth]{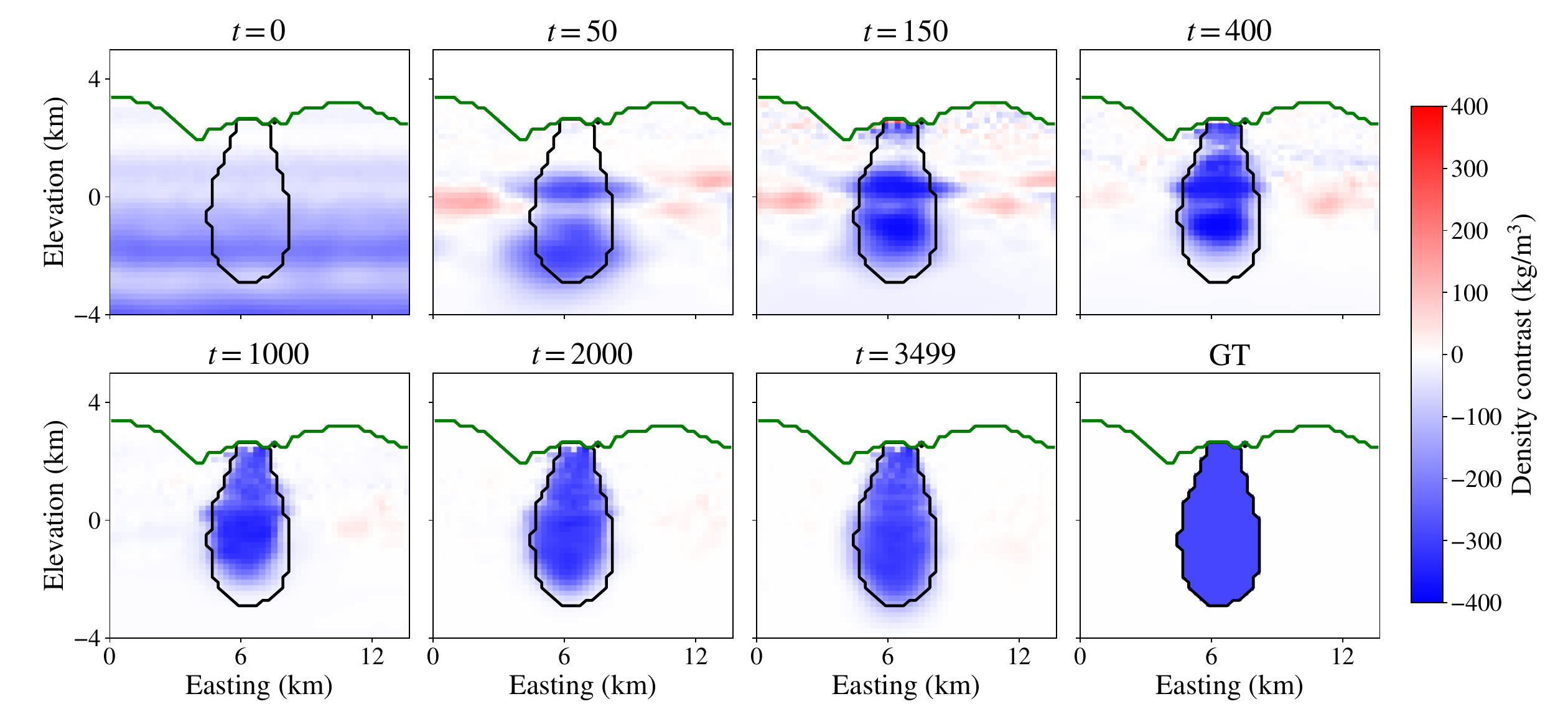}
    \caption{Optimization trajectory of the proposed method for the Cerro Mach\'in synthetic experiment. The reconstruction evolves from a diffuse initial field to a compact and vertically continuous negative anomaly. The ground-truth model (GT) is shown for reference.}
    \label{fig:trajectory}
\end{figure}

The cross-sections in Figure \ref{fig:syntheticslice} confirm the observations from the three-dimensional views. In S1, IRLS fills the interior of the hollow body, while IGI follows parts of the boundary but introduces discontinuities and isolated responses. The proposed method produces a continuous U-shaped anomaly with a visibly reduced, although not completely empty, interior. In S2, the proposed vertical section follows the V-shaped ground-truth contour more closely than the smooth IRLS result and the fragmented IGI reconstruction. In S3, the two depth levels are retained by the proposed method, whereas IRLS merges them into a single diffuse response and IGI introduces discontinuous artifacts. In the Cerro Mach\'in sections, the proposed negative anomaly follows the lateral position and vertical extent of the main ground-truth body more closely than IRLS; it also retains the smaller positive body.

Figure~\ref{fig:responsesynth} shows that the method providing the closest density geometry is not necessarily the one with the smallest data residual. For S1, the receiver-domain RMSE values are 0.062~mGal for IRLS, 0.509~mGal for IGI, and 0.379~mGal for the proposed method. For S2, the corresponding values are 0.035, 0.300, and 0.334~mGal, and for S3 they are 0.066, 0.391, and 0.191~mGal. Thus, IRLS gives the lowest response error in S1--S3 despite its stronger smoothing and loss of internal geometry. In the Cerro Mach\'in case, IGI gives the lowest response error (0.001~mGal), followed by the proposed method (0.071~mGal) and IRLS (0.277~mGal); nevertheless, the IGI density model is more concentrated near the shallow part of the negative anomaly and does not recover its complete depth extent as clearly as the proposed model. These results illustrate the non-uniqueness of the inverse problem: a close gravity fit can coexist with an inaccurate or incomplete density distribution.
\begin{figure}
    \centering
    \includegraphics[width=\columnwidth]{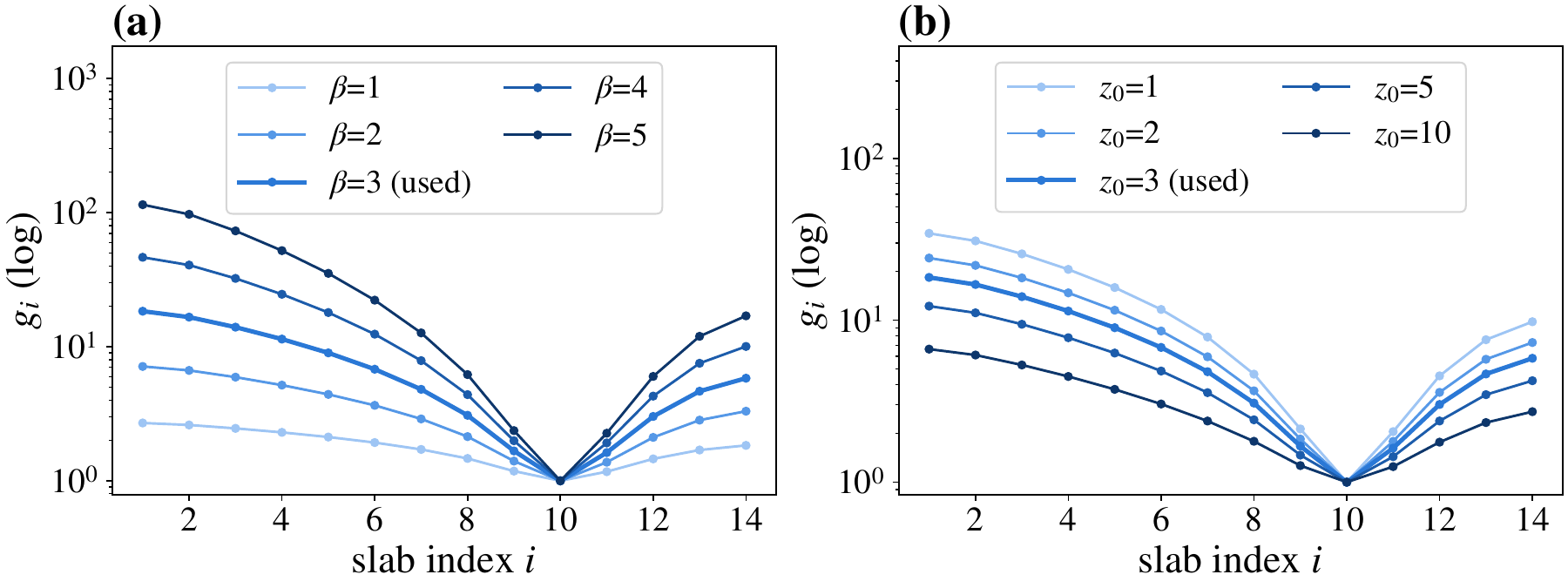}
    \caption{Gain profiles $g_i$ versus slab index $i$ for the Cerro Mach\'in configuration: (a) varying $\beta$ with $z_0=3$ and (b) varying $z_0$ with $\beta=3$. The gains are computed using \eqref{eq:gains}. The selected configuration, $(\beta,z_0)=(3,3)$, is highlighted.}
    \label{fig:gains}
\end{figure}

Figure~\ref{fig:trajectory} shows how the Cerro Mach\'in reconstruction develops during optimization. At $t=0$, the density field is diffuse and dominated by depth-dependent background variation. By $t=50$ and $t=150$, the location of the principal negative anomaly is already visible, but shallow positive artifacts and an extended background remain. Between $t=400$ and $t=1000$, the negative body becomes vertically continuous and the surrounding artifacts are strongly reduced. The later iterations mainly sharpen its lateral concentration and stabilize its depth extent. The final estimate at $t=3499$ reproduces the main location and elongated geometry of the ground-truth body, while retaining smoother boundaries than the reference.

\begin{figure*}[!t]
    \centering
    \includegraphics[width=1\linewidth]{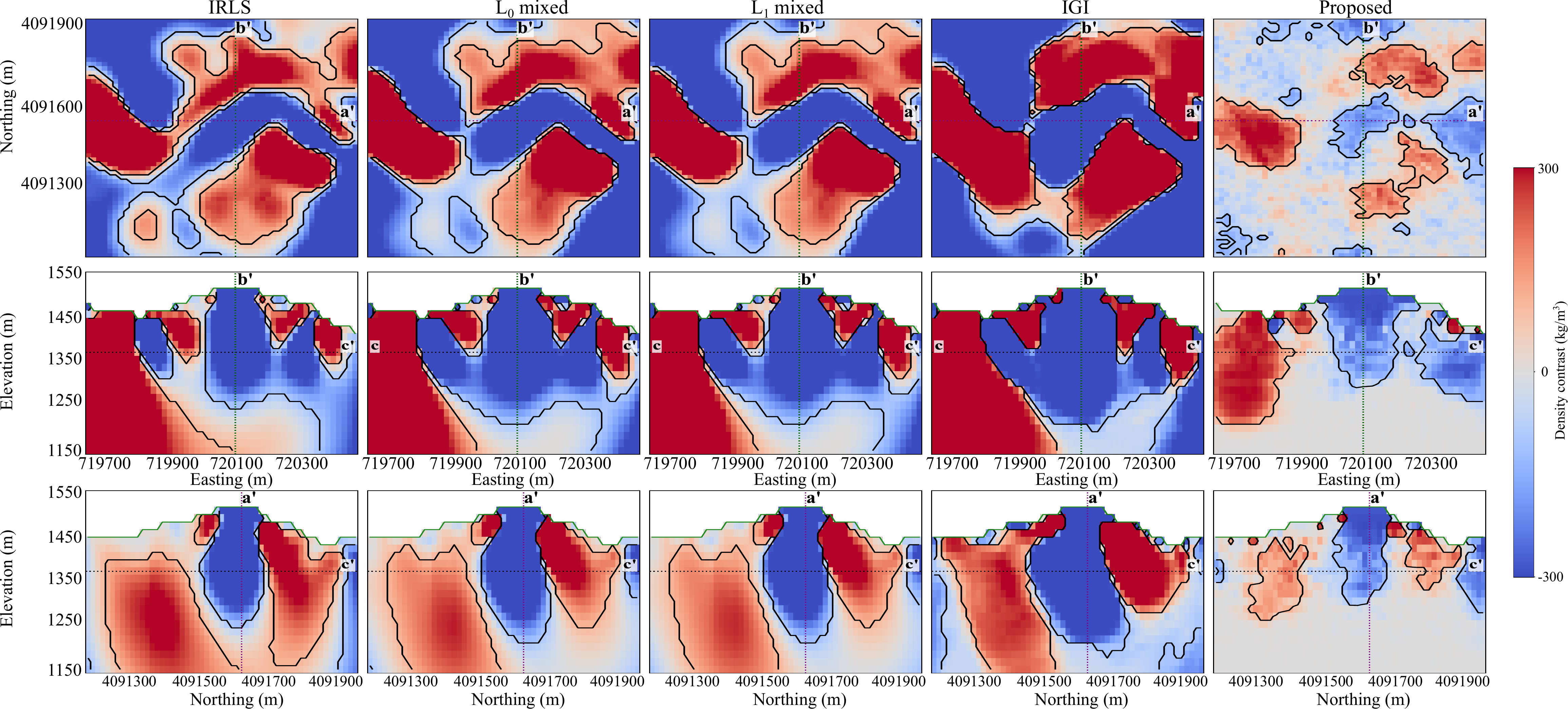}
    \caption{Visual comparison of IRLS, the $L_0$-mixed and $L_1$-mixed methods, IGI, and the proposed method for the Ghareh-Aghaj field data. The first row is a horizontal slice, and the second and third rows are mutually intersecting vertical cross-sections. The dotted lines indicate the locations from which the corresponding sections were extracted.}
    \label{fig:Gharehslices}
\end{figure*}

Figure~\ref{fig:gains} isolates the effect of the two depth-gain parameters for the synthetic Cerro Machín model. Increasing $\beta$ steepens the gain profile and therefore increases the relative capacity assigned to slabs farther from the most sensitive layer. Changing $z_0$ modifies the curvature of the profile, with its strongest effect near the reference layer. For the Cerro Machín mesh, the minimum gain occurs at slab $i=10$, which contains the layer of maximum average sensitivity. The gains increase toward deeper slabs, counteracting the decay of gravity sensitivity with depth. They also increase slightly for slabs located above the topographic surface; however, because these slabs are dominated by inactive cells, this upper branch has little influence on the recovered density model. Thus, Figure~\ref{fig:gains} characterizes the depth weighting imposed specifically in the Cerro Machín synthetic experiment and does not establish a universal optimum for $\beta$ or $z_0$, whose appropriate values depend on the acquisition geometry, topography, sensitivity distribution, and model discretization.

\subsection{Experiment II: Field data}

The field experiment uses the Ghareh-Aghaj potash gravity data described in \cite{sadraeifar2026improved}. The processed data vector used in the inversion contains $N_d=138$ observations. The inversion domain was discretized on a $50\times55\times40$ grid that incorporates the topography, with cell dimensions $\Delta x=14.66875$~m, $\Delta y=13.55909$~m, and $\Delta z=18.64375$~m. The resulting model spans approximately 733.44~m in easting and 745.75~m in both northing and elevation. Because no independent ground-truth density model is available, the recovered structures cannot be ranked by direct model error. Consequently, model-domain PSNR, RMSE, and SSIM are not reported for this experiment. The comparison instead considers the predicted gravity response together with the compactness, continuity, and separation of the recovered anomalies. The hyperparameter configuration used for this experiment, together with the corresponding IGI baseline settings, is summarized in Table~\ref{tab:params_field}.

\begin{table}[!t]
\caption{Hyperparameter configuration for the Ghareh-Aghaj field inversion for the proposed method and the IGI baseline \cite{Li2026ImplicitInversion}. Entries marked --- have no counterpart in the IGI formulation.}
\label{tab:params_field}
\centering
\begin{tabular}{l|c|c}
\hline
\textbf{Parameter} & \textbf{Proposed} & \textbf{IGI baseline} \\
\hline
Grid $(X\times Y\times Z)$             & $50\times55\times40$ & $50\times55\times40$ \\
Observations $N_d$                     & 138 & 138 \\
Slabs $P$                              & 14 & --- \\
Window factor $\kappa$                 & 0.7 & --- \\
Fourier frequencies $L$                & 64 & --- \\
Max. bandwidth $\sigma_{\max}$        & 24 & --- \\
Hidden width                           & 128 & 1024, 1024, 1024, \\
                                       &     & 1024, 512, 256, \\
                                       &     & 64, 16, 8 \\
Hidden layers                          & 4 & 9 \\
Depth exponent $\beta$                 & 4.0 & --- \\
Depth offset $z_0$                     & 3.0 & --- \\
$\lambda_{\mathrm{TV}}$               & $1.5\times10^{-3}$ & --- \\
$\lambda_{L_1}$                    & $4.5\times10^{-3}$ & --- \\
$\lambda_{\mathrm{c}}$                & 0.09 & --- \\
Decay rate $r$                         & 10 & --- \\
Learning rate                          & $10^{-3}$ & $10^{-3}$ \\
Iterations $T$                         & 3500 & 5000 \\
Seed                                   & 0 & 42 \\
\hline
\end{tabular}
\end{table}

Figure~\ref{fig:Gharehslices} shows substantial model variability among solutions that reproduce the same broad gravity pattern. IRLS and the two mixed regularizations recover a large central negative-density region bordered by broad positive-density zones. Their models are spatially continuous but contain extensive regions close to the imposed density limits, and adjacent positive bodies partially merge. The $L_0$-mixed result is slightly sharper than the $L_1$-mixed result, while the latter is smoother. IGI produces a similarly blocky organization, with a large negative central body and strongly saturated positive flanks.
\begin{figure*}[!t]
    \centering
    \includegraphics[width=1\linewidth]{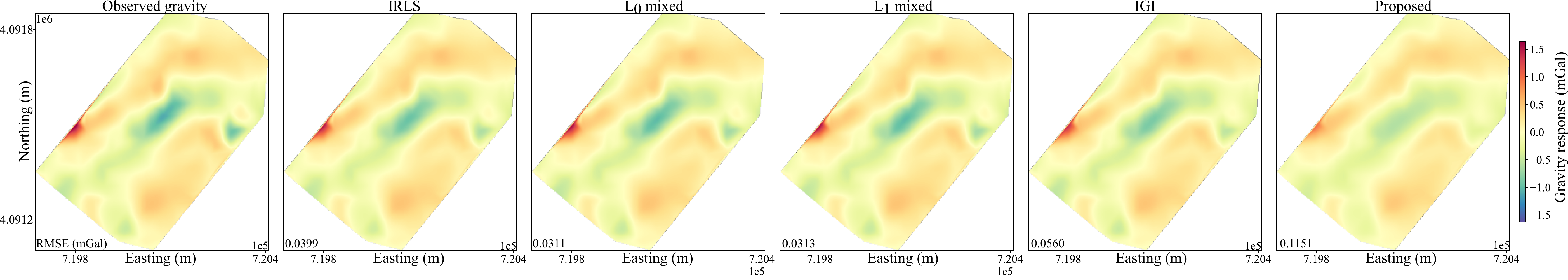}
    \caption{Observed Ghareh-Aghaj gravity response and responses predicted by IRLS, the $L_0$-mixed and $L_1$-mixed methods, IGI, and the proposed method. The number below each predicted response is the receiver-domain RMSE in mGal.}
    \label{fig:responsefield}
\end{figure*}
The proposed reconstruction retains the same first-order arrangement, a central negative anomaly with positive anomalies to the west and east, but distributes substantially less density over the background. In the vertical sections, its main negative body is localized beneath the central part of the survey and remains vertically coherent. The positive bodies are more clearly separated from the negative region than in the baseline models. This compactness is accompanied by lower-amplitude, small-scale variations and less continuity in some peripheral structures. Therefore, the field figure supports the interpretation that the proposed parameterization imposes a stronger compactness and separation prior, but it does not establish that the resulting model is uniquely more accurate.

The response maps in Figure \ref{fig:responsefield} quantify the trade-off observed in the recovered models. The $L_0$-mixed and $L_1$-mixed methods give the smallest RMSE values, 0.0311 and 0.0313~mGal, respectively, followed by IRLS at 0.0399~mGal and IGI at 0.0560~mGal. The proposed method gives an RMSE of 0.1151~mGal. Its predicted response retains the broad central negative trend and the principal positive regions of the observed map, but it attenuates the local extrema and smooths some of the shorter-wavelength variation. The proposed field result should consequently be interpreted as a more strongly regularized solution: it yields a compact, separated density model at the cost of a larger receiver-domain residual. 

\section{Discussion}

The synthetic experiments demonstrate that model-domain accuracy and receiver-domain data fit describe different aspects of gravity inversion. PSNR and model-domain RMSE emphasize voxel-wise agreement with the known density volume, whereas SSIM is more sensitive to spatial organization. Receiver-domain RMSE instead measures how closely a reconstructed model reproduces the gravity observations. Because the inverse problem is non-unique, these measures need not rank the methods in the same order.

Across the four synthetic scenarios, the proposed method obtains the strongest overall model-domain performance, with average values of $25.83$~dB PSNR, $42.18~\mathrm{kg\,m^{-3}}$ RMSE, and $0.704$ SSIM. IRLS gives comparable average PSNR and RMSE, $25.30$~dB and $44.22~\mathrm{kg\,m^{-3}}$, but its average SSIM decreases to $0.534$. This behavior is consistent with its tendency to preserve the broad source location while smoothing boundaries, filling cavities, and merging neighboring bodies. IGI reaches an average SSIM of $0.603$, but its lower PSNR of $22.23$~dB and higher RMSE of $62.85~\mathrm{kg\,m^{-3}}$ reflect the isolated and fragmented structures visible in its reconstructions.

The S1 and S2 cases demonstrate why the metrics must be interpreted jointly. IRLS produces the lowest RMSE and highest PSNR in both scenarios, but its SSIM is lower because the smooth solution loses important internal geometry. In S1, the proposed method raises SSIM from $0.493$ for IRLS to $0.619$ while preserving the hollow structure more clearly. In S2, IGI has a slightly higher SSIM than the proposed method, but its response is discontinuous; visual assessment therefore remains necessary to distinguish structural similarity from spatial coherence. The proposed model provides a more balanced reconstruction, retaining the two limbs without the extensive smoothing of IRLS or the fragmentation of IGI.

The benefits of the depth-aware compositional representation are clearest in S3 and Cerro Mach\'in, where the proposed method gives the best PSNR, RMSE, and SSIM simultaneously. In S3, it improves body separation and raises SSIM to $0.786$. In Cerro Mach\'in, it reduces RMSE from $49.95$ to $35.12~\mathrm{kg\,m^{-3}}$ relative to IRLS and raises SSIM from $0.565$ to $0.873$. These results indicate that slab-specific spectral bandwidths, depth gains, and inter-slab consensus are particularly helpful for models containing multiple anomalies or extended vertical structures.

The receiver-domain results nevertheless show that improved density reconstruction does not always produce the smallest gravity residual. IRLS gives the lowest response RMSE for S1--S3, and IGI gives the lowest response RMSE for Cerro Mach\'in, even when their density estimates are smoother, more fragmented, or less complete at depth. The proposed method therefore acts primarily as a structural prior: it accepts a moderate increase in data residual to favor compactness, continuity, source separation, and depth localization.

The optimization trajectory supports this interpretation in Figure \ref{fig:trajectory}. The large-scale location of the Cerro Mach\'in anomaly is identified early, while the scheduled total-variation and sparsity penalties progressively suppress diffuse background responses. At the same time, the increasing consensus term encourages neighboring slab networks to form a vertically coherent model. The gain profiles show how the representation compensates for depth-dependent sensitivity, but they do not establish universal values for $\beta$ or $z_0$ because their appropriate settings depend on the acquisition geometry, topography, sensitivity distribution, and discretization.

For the field experiment, PSNR, model-domain RMSE, and SSIM cannot be computed because the true density volume is unknown. All methods reproduce the principal polarity pattern of the observed anomaly, but their recovered models differ substantially. The mixed methods, IRLS, and IGI obtain smaller receiver-domain residuals, whereas the proposed method produces a more compact and separated density distribution with a background closer to zero. The field result should therefore be interpreted as an alternative strongly regularized solution rather than as a quantitatively verified improvement in subsurface accuracy.

Several limitations remain. The synthetic hyperparameters were selected using ground-truth model error, which is unavailable in field applications. The field result also shows that the current regularization may trade excessive response fidelity for compactness. Future work should investigate discrepancy-based or data-adaptive hyperparameter selection, uncertainty quantification, sensitivity to noise and initialization, and validation using borehole, geological, seismic, or electromagnetic information. Adaptive regularization schedules may help retain the improvements in PSNR, RMSE, and SSIM while reducing the receiver-domain residual.
\section{Conclusion}

We presented a depth-aware compositional implicit neural representation for 3D gravity inversion. The method represents the density volume using multiple coordinate networks assigned to overlapping depth slabs, together with slab-specific Fourier features, physics-based depth gains, and scheduled regularization, and is optimized directly from gravity observations without requiring paired gravity--density training data. Across the synthetic experiments, the proposed method achieved the best reconstruction metrics on average and produced compact, spatially coherent density models with improved preservation of cavities, separation of nearby bodies, and reconstruction of their vertical extent, indicating that the proposed parameterization provides an effective structural prior for mitigating the depth ambiguity inherent in gravity inversion. In the Ghareh-Aghaj field experiment, although the stronger structural regularization resulted in a larger receiver-domain residual than the baselines, it yielded a more localized and vertically coherent subsurface interpretation. Overall, the results demonstrate that the proposed depth-aware representation offers a promising unsupervised alternative for recovering geologically interpretable density models in ill-posed 3D gravity inversion problems.

\section{Acknowledgment}

This work was funded by the Agencia Nacional de Hidrocarburos (ANH) and the Ministerio de Ciencia, Tecnolog\'ia e Innovaci\'on (MINCIENCIAS), under contract 045-2025.  Also, we thank the support of the Vicerrector\'ia de Investigaci\'on y Extensi\'on from Universidad Industrial de Santander under Project 4619. ChatGPT was used to improve the grammar and readability of the manuscript. The authors reviewed and verified all AI-assisted content and remain fully responsible for the final work.

\section{Data and materials availability}
The data sets and code used in this study are publicly available for reproducibility and further research at \url{https://github.com/leonsuarez24/Depth-gravity-inversion}.

\bibliographystyle{IEEEtran}
\bibliography{references}

\phantomsection

\begin{IEEEbiography}[{\includegraphics[width=1in,height=1.25in,clip,keepaspectratio]{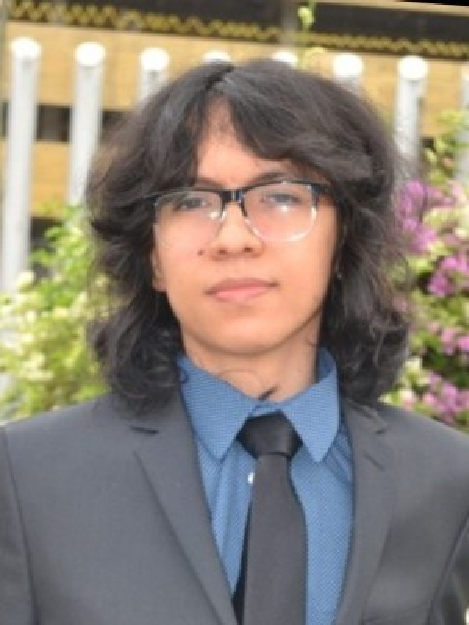}}]{Leon Suarez-Rodriguez}received a B.S.E. degree in Civil Engineering from Universidad Industrial de Santander, Bucaramanga, Colombia, in 2023, and a Master's degree in Systems Engineering in 2025 from the same institution. He is currently pursuing a Ph.D. in Computer Science at Universidad Industrial de Santander. His research interests include inverse problems and deep learning applications in geosciences and computational imaging.
\end{IEEEbiography}
\vspace{1em}

\begin{IEEEbiography}[{\includegraphics[width=1in,height=1.25in,clip,keepaspectratio]{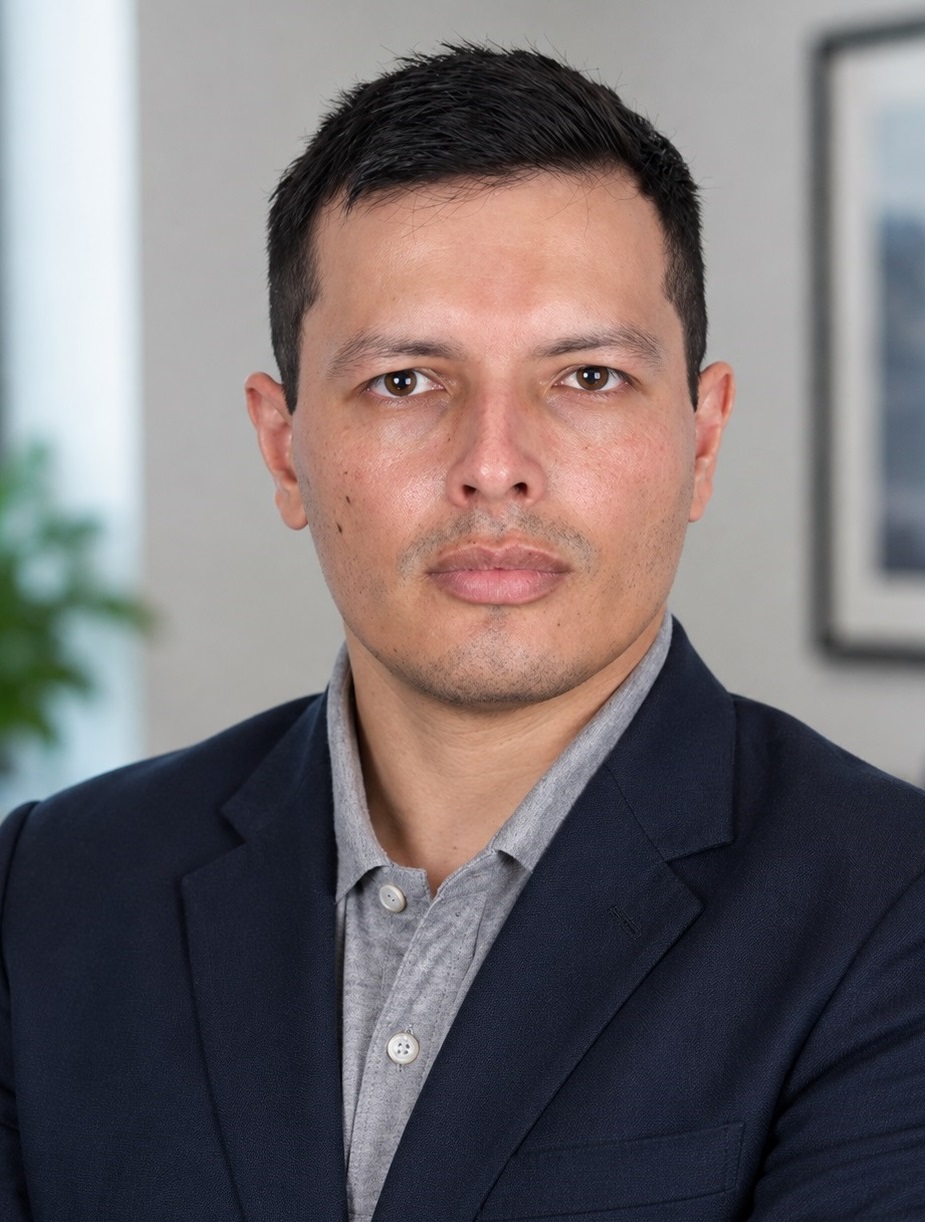}}]{Paul Goyes-Pe\~nafiel}
received the B.Sc. in geology from the Universidad Industrial de Santander, Bucaramanga, Colombia, and M.Sc. in geophysics from the Perm State University, Perm, Russia, in 2009 and 2018, respectively. He is currently a Ph.D. candidate in Computer Science with the Universidad Industrial de Santander. He has been a researcher in the hydrocarbon industry, focusing on applied geophysics for both shallow and deep exploration. His research interests are in inverse theory and applications in geophysics, seismic acquisition and processing, potential-EM methods, and deep learning applications in geoscience.
\end{IEEEbiography}
\vspace{1em}

\begin{IEEEbiography}[{\includegraphics[width=1in,height=1.25in,clip,keepaspectratio]{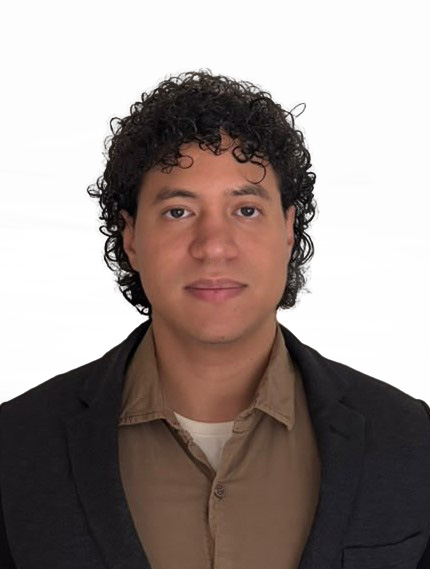}}]{Javier Torres-Quintero}
received the B.Sc. degree in Systems Engineering in 2024 and is currently pursuing the M.Sc. degree in Systems Engineering and Informatics and the Ph.D. degree in Computer Science at the Universidad Industrial de Santander, Bucaramanga, Colombia. His research interests include deep learning, computational imaging, seismic processing, hyperspectral image analysis, and signal processing applications in geoscience.
\end{IEEEbiography}
\vspace{1em}

\begin{IEEEbiography}[{\includegraphics[width=1in,height=1.2in,clip,keepaspectratio]{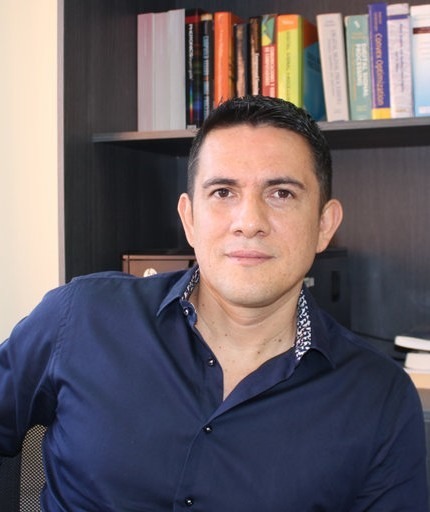}}]{Henry Arguello}
 received his Ph.D. degree from the Electrical and Computer Engineering Department at the University of Delaware in 2013. He is currently a titular professor in the Systems Engineering Department, Universidad Industrial de Santander, Bucaramanga, Santander 680002, Colombia. He was a visiting Professor at Stanford funded by Fulbright.  He is an Associate Editor for the IEEE Transactions on Computational Imaging. He was the Co-Chair and the Technical Co-Chair of several international conferences and workshops. His research interests include computational imaging techniques, high-dimensional signal coding and processing, and optical design. He is a Senior Member of IEEE\end{IEEEbiography}

\end{document}